\documentclass[10pt,twocolumn,letterpaper]{article}

\usepackage{iccv}
\usepackage{times}
\usepackage{epsfig}
\usepackage{graphicx}
\usepackage{amsmath}
\usepackage{amssymb}
\usepackage{graphicx}
\usepackage{amsmath}
\usepackage{footmisc}
\usepackage{amssymb}
\usepackage{booktabs}
\usepackage{bbding}
\usepackage{multirow}
\usepackage[table,xcdraw]{xcolor}
\usepackage{verbatim}
 \usepackage{graphicx}
 \usepackage{adjustbox} 
 \usepackage{array}
\usepackage[accsupp]{axessibility}
\usepackage{eso-pic}

\usepackage[breaklinks=true,bookmarks=false]{hyperref}

\iccvfinalcopy % *** Uncomment this line for the final submission

\def\iccvPaperID{1674} % *** Enter the ICCV Paper ID here

\ificcvfinal\fi

\begin{document}

%%%%%%%%% TITLE
\title{MDCS: More Diverse Experts with Consistency Self-distillation for Long-tailed Recognition}

\author{Qihao Zhao$^{1, 2, }$\thanks{Equal contribution.} , Chen Jiang$^{1, }$\footnotemark[1] , Wei Hu$^1$, Fan Zhang$^{1}\thanks{The Corresponding author is with the College of Information Science and Technology and the Interdisciplinary Research Center for Artificial Intelligence, Beijing University of Chemical Technology, China}$\:, Jun Liu$^2$\\
$^1$ Beijing University of Chemical Technology, China.\\
$^2$ Singapore University of Technology and Design, Singapore\\
{\tt\small \{zhaoqh,jiangchen,huwei,zhangf\}@mail.buct.edu.cn}, 
{\tt\small jun\_liu@sutd.edu.sg}
}

\maketitle
\ificcvfinal\thispagestyle{empty}\fi

\begin{abstract}
Recently, multi-expert methods have led to significant improvements in long-tail recognition (LTR). We summarize two aspects that need further enhancement to contribute to LTR boosting: (1) More diverse experts; (2) Lower model variance. However, the previous methods didn't handle them well. To this end, we propose More Diverse experts with Consistency Self-distillation (MDCS) to bridge the gap left by earlier methods. Our MDCS approach consists of two core components: Diversity Loss (DL) and Consistency Self-distillation (CS). In detail, DL promotes diversity among experts by controlling their focus on different categories. To reduce the model variance, we employ KL divergence to distill the richer knowledge of weakly augmented instances for the experts' self-distillation. In particular, we design Confident Instance Sampling (CIS) to select the correctly classified instances for CS to avoid biased/noisy knowledge. In the analysis and ablation study, we demonstrate that our method compared with previous work can effectively increase the diversity of experts, significantly reduce the variance of the model, and improve recognition accuracy. Moreover, the roles of our DL and CS are mutually reinforcing and coupled: the diversity of experts benefits from the CS, and the CS cannot achieve remarkable results without the DL. Experiments show our MDCS outperforms the state-of-the-art by 1\% $\sim$ 2\% on five popular long-tailed benchmarks.%, including CIFAR10-LT, CIFAR100-LT, ImageNet-LT, Places-LT, and iNaturalist 2018.% The code is available at \url{https://github.com/fistyee/MDCS}

\end{abstract}

%%%%%%%%% BODY TEXT
\section{Introduction}
\label{sec:intro}

\begin{figure}[t]
\centering
\includegraphics[width=1\columnwidth]{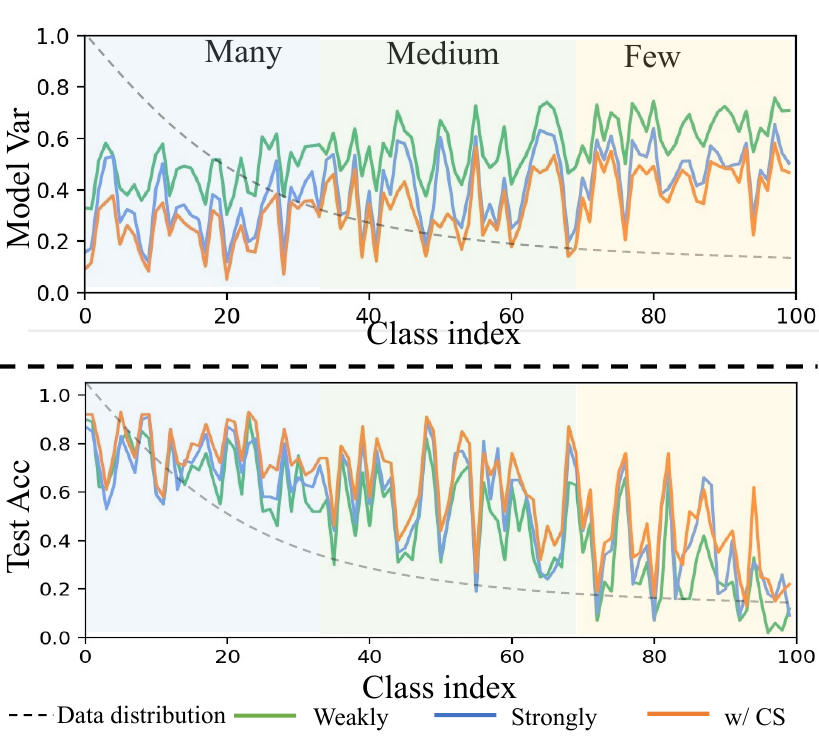}
\caption{We evaluate a ResNet-32 model based on Balanced Softmax \cite{ren2020balanced, reweightmenon2020long} with weakly/strongly augmentation. All experiments were performed with an Imbalanced Factor (IF) of 100 on the CIFAR100-LT dataset. \textbf{Top}: model variance \cite{wang2020longRIDE}. The model trained with weakly augmented instances has a higher variance, whereas the model trained with strongly augmented instances is better than weak augmentation. \textbf{Bottom}: Test accuracy. In the case of training on weakly/strongly augmented instances, the model supervised with one-hot labels presents lower accuracy. In contrast, our CS transfers richer knowledge from weakly augmented instances, preventing the model from overfitting instances as well as reducing the model variance and improving recognition accuracy. It indicates that the prediction of the weakly augmented model could provide richer supervision knowledge to the strongly augmented instances better than its one-hot label.}

\label{figmodelvar}
\end{figure}

Deep learning has achieved remarkable progress in a range of computer vision (CV) tasks, such as image recognition \cite{he2016deep, vit, zhao2022mixpro, liu2021swin,zhao2021p}, object detection \cite{sun2021sparse,dai2021dynamic}, and action recognition \cite{sun2022human}. Despite advances in deep technologies and computational capability, great success is also highly dependent on well-designed large datasets such as ImageNet \cite{imagenet} and Places \cite{zhou2017places}, where each category has sufficient and roughly balanced training samples. However, real-world data tends to be long-tailed over semantic categories \cite{zhang2021longtailsurvey}: a few categories contain many instances (called head categories), while most categories contain only a few instances (called tail categories). Long-tailed recognition (LTR) is challenging because it needs to deal not only with the numerous small data learning problems of the tail categories but also with the extremely unbalanced classification of all categories. Deep models trained with such long-tailed data are usually biased toward head categories on balanced testing data and perform poorly on tail categories.

To address this challenge, many approaches have explored long-tail recognition in order to learn well-performing models from long-tailed data, such as class re-balancing/re-weighting \cite{oversamplebuda2018systematic,oversamplebyrd2019effect,LTkhan2017cost,focalloss,reweightcui2019class,reweightcao2019learning,reweightxie2019intriguing,reweightmenon2020long}, decoupling learning \cite{LTkang2019decoupling} and contrastive learning \cite{yang2020rethinking,kang2020exploring,wang2021contrastive,zhu2022balanced,cui2021parametric}. Recently, long-tailed recognition methods employing multi-expert ensemble learning \cite{xiang2020learning, wang2020longRIDE,cai2021ace,zhang2022SADE,li2022nested} have achieved state-of-the-art (SOTA) performance. We summarize two key aspects of these approaches that need further improvement for boosting LTR. (1) Diverse experts experts focus on different aspects, maximizing the expertise of each \cite{cai2021ace, wang2020longRIDE}. More diversity can support experts in improving LTR. (2) There is a heavy model variance in the prediction of the model, especially for the tail category. So, reducing model variance is essential for LTR. Previous multi-expert methods \cite{xiang2020learning, wang2020longRIDE,cai2021ace,zhang2022SADE,li2022nested} focused on the above two aspects but did not handle them well. RIDE \cite{wang2020longRIDE} utilizes a loss to moderate diversity, yet individual experts focus primarily on head categories. ACE and SADE \cite{cai2021ace, wang2020longRIDE, zhang2022SADE} focus on the diverse experts, which learn classification knowledge from sub-categories or dominant categories. However, The "tail category experts" of these methods can greatly suppress head category performance while focusing on the tail categories. Furthermore, these multi-expert methods all employ an ensemble method to reduce the final variance while ignoring the variance of each expert. Among them, NCL \cite{li2022nested} introduces strong data augmentation \cite{cubuk2020randaugment} that provides a better generalization of the model. However, there is still a high risk of model variance in its one-hot label supervision for strongly augmented instances. To this end, we design a novel method, namely \textbf{M}ore \textbf{D}iverse experts with \textbf{C}onsistency \textbf{S}elf-distillation (MDCS), for long-tailed recognition.

Our proposed MDCS contains two key components, Diversity Loss (DL) and Consistency Self-distillation (CS). Our DL contains an adjustable distribution weight, to cater to the diversity of each expert. By adjusting the distribution weight, each expert tends to recognize different categories, such as Many-shot categories, Medium-shot categories, and Few-shot categories. It is a simple yet effective method for increasing diversity and significantly improving recognition accuracy over previous methods (discussed in Sec. \ref{Secanalysis}). To reduce the model variance and avoid model overfitting instances, we look forward to providing each expert with a richer form of supervision when learning strongly augmented samples. The label-smoothing regularization \cite{LS1, LS2} is a straightforward way, and further MiSLAS \cite{Zhong_2021_CVPR} proposes label-aware smoothing for long-tailed recognition. However, the proportion of label-smoothing assignments of these methods is still instance-agnostic, and more reasonable label assignment principles remain to be explored. To this end, we design CS for each expert, which distills richer instance knowledge from predictions of weakly augmented data to regularize strongly augmented instances. Especially for a mini-batch instance, we propose Confident Instance Sampling (CIS) to select the correctly classified instances for consistency self-distillation. In this way, our proposed CIS can prevent CS from introducing biased/noisy knowledge. As illustrated in Fig. \ref{figmodelvar}, the model trained with strong augmentation method \cite{cubuk2020randaugment} could reduce the model variance \cite{wang2020longRIDE} compared with the model trained with weakly augmented instances (e.g., flipped, cropped). However, our CS trains the model on strongly augmented instances and is supervised by ”soft labels” from the predictions of weakly augmented instances, leading to lower model variance and higher recognition accuracy. These "soft labels", produced by prediction on weakly augmented representation, contain more knowledge than their one-hot labels. In addition, the roles of our DL and CS are mutually reinforcing and coupled: (1) Our CS is designed for each expert, which increases the diversity and recognition accuracy of a single expert, and ultimately benefits the ensemble model. (2) Without the DL, the CS cannot achieve remarkable results as the model is  biased towards head categories (discussed in Sec. \ref{Ablation_Study}). 

In the experiments, our proposed MDCS model outperforms state-of-the-art (SOTA) methods by a significant margin on five commonly used benchmark datasets. For instance, on CIFAR-100-LT with an imbalance factor of 100, our approach achieves an accuracy of 56.1\%. Similarly, on ImageNet-LT with ResNeXt-50, our model achieves an accuracy of 61.8\%, while on iNaturalist 2018 with ResNet-50, we achieve an accuracy of 75.6\%.

%------------------------------------------------------------------------
\section{Related Work}
\label{sec:Related Work}

{\bf Long-tailed Visual Recognition.} \quad Conventional methods to alleviate the long-tailed problem are to design re-balancing paradigms that consist of re-sampling and re-weighting. Re-sampling methods, which over-sample tail classes or under-sample head classes, aim to achieve a more balanced data distribution. Re-sampling by simply over-sampling minority classes \cite{oversamplebuda2018systematic,oversamplebyrd2019effect, Park_2022_CVPR} and under-sampling by abandoning data for dominant classes \cite{undersamplejapkowicz2002class,undersamplehe2009learning,oversamplebuda2018systematic}. However, over-sampling duplicated tailed samples might lead to the over-fitting of minority classes \cite{oversamplebuda2018systematic}. Under-sampling potentially loses head class information and certainly impairs the generalization ability of the DNNs. Re-weighting methods \cite{LTkhan2017cost,focalloss,reweightcui2019class,reweightcao2019learning,reweightxie2019intriguing,reweightmenon2020long,reweightren2020balanced, xu2022constructing} assign weights to different classes by loss modification or logits adjustment. However, some researchers observed that re-balancing methods will hurt representation learning, and decoupling representation with classifiers will lead to better features. Therefore, two-stage learning was proposed which first trains the model with original data and then fine-tunes the classifier with class-balanced data \cite{decoupling}. %BBN\cite{LTzhou2020bbn} handles both representation and classifier simultaneously through a two-brunch network with a cumulative learning strategy. 
Transfer learning is another way to tackle the long-tailed problem, aiming to transfer knowledge learned from majority classes to minority classes. But knowledge transfer methods often need carefully designed structures such as memory bank \cite{liu2019large,9577321, Long_2022_CVPR}. More recently, many works try to improve the performance of long-tailed visual recognition by using contrastive learning (CL) strategy \cite{yang2020rethinking,kang2020exploring,wang2021contrastive,zhu2022balanced,cui2021parametric,li2022targeted}. For example, PaCo \cite{cui2021parametric} introduces a set of class-wise learnable center to overcome bias on high-frequency classes of basic supervised contrastive learning (SCL). %BCL \cite{zhu2022balanced} further improves the hybrid network structure through BCL loss. 

\begin{figure*}[t]
\centering
\includegraphics[width=2\columnwidth]{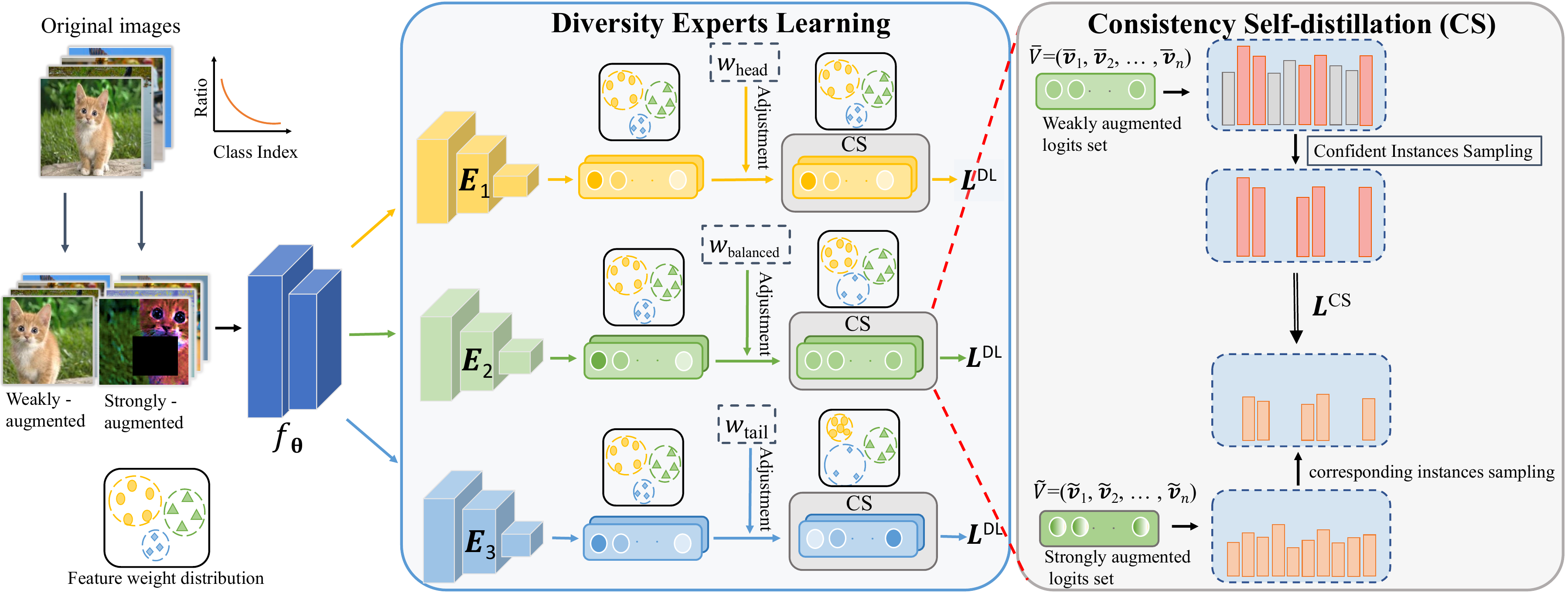}
\caption{Our method consists of two core components: (i) Diversity Loss (DL), which trains diversity experts; and (ii) Consistency Self-distillation (CS), which reduces the variance of each expert. Firstly, original images with weak and strong augmentation as inputs for shared backbone experts. Secondly, our DL controls experts' focus on different categories with an adjustable distribution weight (\emph{W}) to learn different feature weight distributions. Finally, our proposed Consistency Self-distillation distills the richer knowledge from the prediction of weakly augmented instances to address overfitting and reduce model variance. Meanwhile, our proposed Confident Instances Sampling prevents CS from introducing biased knowledge.}

\label{figFramework}
%\vspace{-10px}
\end{figure*}

Ensemble-based methods, which use multiple experts with aggregation methods, are receiving more and more attention due to their effectiveness on long-tailed recognition. LFME \cite{xiang2020learning} trains different experts with different parts of the dataset and distills the knowledge from these experts to a student model. RIDE \cite{wang2020longRIDE} optimizes experts jointly with distribution-aware diversity loss and trains a router to handle hard samples. SADE \cite{zhang2021test} proposed test-time experts aggregating method to handle unknown test class distributions. Recently proposed NCL \cite{li2022nested} uses mutual distillation allowing every expert to learn knowledge from others. They still have shortcomings in terms of expert diversity and model variance.

{\bf Knowledge Distillation.} \quad Knowledge distillation(KD) \cite{buciluǎ2006model,hinton2015distilling, passalis2018learning} was proposed for the purpose of model compression by transferring knowledge learned from a powerful teacher model to a student model. KD is performed by supervising the student model with soft labels generated by the teacher model, which also provides better generalization to the student model. KD has gradually evolved from an offline process \cite{peng2019correlation, hinton2015distilling,passalis2018learning}, where the teacher model has trained ahead of time, to an online process \cite{chen2020online,guo2020online, zhang2018deep}, where teacher model and student model are trained simultaneously. Unlike offline or online KD, self-distillation \cite{zhang2019your} assumes that one model can be its own teacher, where the teacher model and student model are identical. %The effectiveness of knowledge distillation has been tested in many domains \cite{li2017learning,pmlr-v80-furlanello18a,bagherinezhad2018label,liu2019structured} beyond model compression. 

{\bf Consistency regularization.} Consistency regularization has played a very important role in semi-supervised learning, which was first proposed by Bachman \cite{bachman2014learning} and  popularized by Sajjadi \cite{sajjadi2016regularization} and Laine \cite{laine2016temporal}. Consistency regularization utilizes unlabeled data by assuming the model should output the same result when the inputs are similar. Specifically, given two different formations of perturbation of a training sample, the gap in the output is treated as a loss to train the model. There are various ways to generate perturbed input \cite{miyato2018virtual,tarvainen2017mean}. A common method is employing two different formations of data augmentation on the same image \cite{sohn2020fixmatch}. FixMatch \cite{sohn2020fixmatch} computes an artificial label for each unlabeled sample by computing the model’s predicted class distribution given a weakly-augmented version. Unlike the above methods, our proposed consistency self-distillation first designs without extra hyper-parameters, combines consistency mechanisms that transfer the richer knowledge of weakly augmented instances to provide more supervision, and employs confidence instance sampling to remove biased/noisy knowledge. Benefiting from these well-designed components, our method effectively reduces the model variance and improves generalization ability.

\section{Method}
\label{secMethod}

The proposed MDCS consists of two parts, Diversity Loss(DL) and Consistency Self-distillation(CS), aiming to improve the diversity of experts and reduce model variance, respectively. In the following part, we first introduce the preliminaries of long-tailed recognition. Then, we elaborate on our proposed DL and CS. Finally, we show the overall loss of the training process.

\subsection{Preliminaries}
Long-tail identification attempts to learn a well-represented classification model from a training dataset with a long-tailed class distribution. Formally, let $\mathbb{D}_s = \{(x_i, y_i)|1 \le i \le n_s\}$ be a training set, which $x_i$ is the $i$-th training sample and $y_i \in\{0,1\}^C$ is its corresponding one-hot label over C classes. The test set $\mathbb{D}_t = \{(x_i, y_i)|1 \le i \le n_t\}$ is defined in the similar way. Let $n_j$ denote the number of training samples for
class $j$, and let $N = \sum_{j=1}^{C}{n_j}$ be the total number of training samples. Without loss of generality, we assume that the classes are decreasingly ordered, i.e., if $i  $\textless$  j$, then $n_i \ge n_j$. Additionally, an imbalanced dataset has significant differences in the class instance numbers, $n_i \gg n_j$. %Various loss functions have been developed for long-tailed recognition $f(x_i; \theta) \mapsto \vec{v_i}$ on the training dataset, where $f(x_i; \theta)$ is implemented by a DNN with parameter $\theta$. %Among them, the most popular one is the balanced softmax cross-entropy (BSCE) loss:

%\begin{equation}\label{equCELoss}
%  \mathcal{L}_{CE}=-\sum\limits_{m=1}^C y_mlog(p_m), 
%\end{equation}
%where the conditional probability $p_m$ can be expressed as

%\begin{equation}\label{equSoftmax}
%  p_m = \frac{e^{\eta_m}}{\sum_{j=1}^{C}{e^{\eta_j}}}.
%\end{equation}
%The softmax function maps a model’s class-m output $\eta_m = {\vec{W}_{m}^T\vec{v}_{m}+\vec{b}_{m}}$ to the $p_m$. The $\vec{W}$ denotes the classifier weight matrix and $\vec{b}$ denotes the bias term, and the output of the  softmax function is $\vec{P} = (p_1, \ldots , p_C)$.

\subsection{Diversity Loss}

For training diversity experts, one intuitive approach is to train different experts to focus on different sub-categories. We propose our Diversity Softmax defined as:

\begin{equation}\label{ExtenedBalancedSoftmax}
    p(x;\theta) = \frac{n_k^\lambda exp({v^k})}{\sum_{c=1}^{C}n_c^\lambda  exp({v^c})}, \lambda \in (-\infty, \infty),
\end{equation}

where $v^k$ is the class-k output of the model $f(x_i; \theta)$ with parameter $\theta$, and $n_k$ is the number of training samples for category $k$. The diversity softmax function maps a model’s class-k output $v^k$ to the probability $p(x;\theta)$. More importantly, the introduced $\lambda$ acts as a weight distribution parameter for logit adjustment. %In prior arts \cite{ren2020balanced, reweightmenon2020long}, the $\lambda$ is set to $(0,1)$, which can weaken the influence of long-tailed distribution. Intuitively, it has the ability to simulate any weight distribution, so we extend it to $(-\infty, \infty)$ as our diversity softmax. 
Fortunately, in our experiments, we discover that it has the effect of generating a reversed weight of long-tailed distribution when $\lambda > 1$. Then, we can employ it to train an expert to improve the accuracy of minority categories with original long-tailed data distribution. Similarly, when $\lambda < 0$, it has the effect of aggravating the imbalance of original long-tailed data, which makes the expert pay more attention to the head categories. When the $\lambda$ is set to $(0,1)$, it can weaken the influence of long-tailed distribution \cite{ren2020balanced, reweightmenon2020long}. Notably, the aggravation is sensible because our intention is to improve the diversity of all experts in \textbf{all} categories. The experiments in Sec. \ref{Ablation_Study} also demonstrate the effect of $\lambda$ for simulating weight distribution. 

With diversity softmax, we propose our Diversity Loss (DL) for diversity experts learning. The DL is defined as: 
\begin{equation}\label{diversityLoss}
  \mathcal{L}^{DL} = \frac{1}{\left\| \mathbb{D} \right\|} \sum\limits_{x_i\in \mathbb{D}}-y_i\log \sigma (f(x_i; \theta)+w) ,
\end{equation}
where the $\sigma(\cdot)$ is the standard softmax function and w is: 
\begin{equation}\label{diversityLoss2}
  w = \lambda \log N^C , 
\end{equation}
where $N^C$ is a list consisting of the number of training samples for each category. In the default setting, we employ DL for training three experts, namely $E_{1}$, $ E_{2}$, and $E_{3}$, focusing on Many-shot classes, Medium-shot classes, and Few-shot classes respectively. Visually, Fig. \ref{figFramework} illustrates a multi-expert model with a shared backbone $f_\theta$ and three experts trained with DL. The only difference between the Diverse Loss for different experts is the distribution weight $w$, such as  $w_{head}$ for $E_{1}$, $w_{balance}$ for $ E_{2}$, and $w_{tail}$ for $E_{3}$. In general, the structure of our diversity experts learning can have any number of experts $E_\mu(\mu=1,2,3,...)$. The effect of the number of experts is shown in the later section \ref{Ablation_Study}.  %Our proposed Dl differs from  in that we reasonably extend the range of hyperparameters $\lambda$ to any values as Balanced Softmax aims to weaken the long-tailed distribution while we intend to simulate diversity distribution.

\subsection{Consistency Self-distillation}

\textbf{Overall view.} In this section, we propose an elegant Consistency Self-distillation (CS) approach to tackle the model variance problem. Our CS method distills richer knowledge from a normal image to a distorted version of the same image. As demonstrated in the left part of Fig. \ref{figFramework}, we first construct an original image $x_i$ to two different views denoted as $\overline{x}_i$ and $\tilde{x}_i$ by a weak augmentation (e.g. crop, flip) and a strong argumentation (e.g. RandAug \cite{cubuk2020randaugment}). For different Expert $E_\mu$, we employ diversity softmax to conduct  probabilities $\overline{p}(x_i; \theta)$ and $\tilde{p}(x_i; \theta)$ for given $(\overline{x}_i,\tilde{x}_i)$:

\begin{equation}
\begin{split}
    \overline{p}(x_i; \theta) = \frac{n_k^\lambda exp({\overline{v}_i^k}/T)}{\sum_{c=1}^{C}n_c^\lambda  exp({\overline{v}_i^c} / T)} ,
    \end{split}  
\end{equation}

\begin{equation}
\begin{split}
    \tilde{p}(x; \theta) = \frac{n_k^\lambda exp({\tilde{v}_i^k}/T)}{\sum_{c=1}^{C}n_c^\lambda  exp({\tilde{v}_i^c} / T)} ,
\end{split}   
\end{equation}
where T is a temperature (a higher T produces a softer probability distribution over categories\cite{hinton2015distilling}). Then, our proposed CS employs the Kullback-Leibler(KL) divergence to perform self-distillation for an instance, which can be formulated as:
\begin{equation}
\begin{split}
    \mathcal{L}^{CS} = KL(p(\overline{x}_i;\theta) || p(\tilde{x}_i; \theta)).
\end{split}                     
\end{equation}

\textbf{Confident Instance Sampling}
As our diversity experts specialize in certain categories and may perform poorly in other categories. To prevent CS distills from all instances introducing biased knowledge, we only distill the instances which are correctly classified. Thus, we define Confident Instance set contain all correctly classified instances as:
\begin{equation}
\begin{split}
    \mathbb{D}_{CI} = \{x_i \in \mathbb{D} | argmax(p(\overline{x}_i;\theta)) == y_i \}, 
\end{split}                     
\end{equation}
where $y_i$ is the ground-truth label of instance $x_i$. %The ablation study in section \ref{Ablation_Study} shows the effect of our proposed CIS. 
Furthermore, we re-formulate the loss of CS with CIS  as:
\begin{equation}
\begin{split}
    \mathcal{L}^{CS}= \frac{1}{\left\| \mathbb{D}_{CI} \right\|}\sum_{x_i\in \mathbb{D}_{CI}} KL(p(\overline{x}_i;\theta) || p(\tilde{x}_i; \theta)) 
\end{split}                     
\end{equation}

\subsection{Model Training}
The overall loss in our proposed MDCS consists of two parts, the Loss $\mathcal{L}^{DL}$ of diversity loss and the Loss $\mathcal{L}^{CS}$ of consistency self-distillation with CIS. Finally, we denote the set of Expert as $E$ and formulate the overall loss as:
\begin{equation}
\begin{split}
    \mathcal{L}_{all} = \sum_{E_\mu \in E}(\mathcal{L}^{DL}_{\mu} + \alpha\mathcal{L}^{CS}_{\mu})
\end{split}                     
\end{equation}
where $\alpha$ is a hyperparameter to adjust the weight of Consistency Self-distillation. In addition, we conduct the effect of parameter $\alpha$ in Sec. \ref{Ablation_Study}.

\section{Method Analysis}
\label{Secanalysis}
\subsection{More Diverse Experts}
%As shown in Table \ref{Diversity Factor}, the Diversity Factor of Our MDCS is significantly bigger than SADE \cite{zhang2022SADE} and RIDE \cite{wang2020longRIDE}, which demonstrates that the diversity of our method is much greater.

%\begin{table}[!htb]
%\centering
%\begin{tabular}{cc|ccc}
%\multicolumn{2}{c|}{Model}    

%& MDCS(ours) & SADE   & RIDE   \\ \hline

%\multicolumn{1}{c|}{\multirow{4}{*}{DF}} & Many   & 81.8\%     & 75.5\% & 75.8\% \\
%\multicolumn{1}{c|}{}                    & Medium & 63.2\%     & 57.0\% & 61.5\% \\
%\multicolumn{1}{c|}{}                    & Few    & 60.3\%     & 46.5\% & 37.5\% \\
%\multicolumn{1}{c|}{}                    & All    & 60.3\%     & 52.1\% & 53.1\% \\ \hline

%\end{tabular}
 %   \caption{Diversity Factor}
%\end{table}

\textbf{Definition of diversity.}
According to our empirical analysis, more diverse experts could contribute to the improvement of long-tailed recognition. However, the previous works \cite{zhang2022SADE, wang2020longRIDE} don't present a measure of diversity. Here, we propose a measure called the diversity factor ($\sigma$), defined for a model containing $M$ experts as:

\begin{equation}\label{Diversity Factor}
    \sigma = \bigcup_{\mu=1}^{M}
     S_\mu
\end{equation}
where $S_\mu$ is all the correctly classified samples in the test set by Expert $E_u$. The $S_\mu$ can define as:

\begin{equation}\label{Correct samples}
    S_\mu = \{argmax(p(x_i;\theta_\mu)) == y_i, (x_i,y_i) \in \mathbb{D}_t \}
\end{equation}
The bigger $\sigma$ represents greater diversity for the ensemble model.

\begin{table}[!htb]
    \centering

\renewcommand\arraystretch{1.8}
  \scalebox{0.75}{
   {

    \begin{tabular}{@{}cp{0.6cm}p{0.6cm}p{0.6cm}p{0.6cm}cp{0.6cm}p{0.6cm}p{0.6cm}p{0.6cm}@{}}
    \toprule
                            & \multicolumn{9}{c}{RIDE \cite{wang2020longRIDE}}                                             \\ \cline{2-10} 
                            & \multicolumn{4}{c}{ImageNet-LT} &  & \multicolumn{4}{c}{CIFAR100-LT} \\ \cline{2-5} \cline{7-10} 
    \multirow{-3}{*}{Model} & Many   & Med.   & Few   & All   &  & Many   & Med.   & Few   & All   \\ \hline
    E1 Acc                     & 64.3   &  49.0      & 31.9      &52.6       &  & 63.5        &44.8        &20.3       &44.0       \\
    E2 Acc                     & 64.7   &49.4       &31.2       &52.8       &  &63.1        &44.7        &20.2       &43.8       \\
    E3 Acc                     & 64.3   &48.9        &31.8       &52.5       &  &63.9        &45.1        &20.5       &44.3       \\
    \rowcolor[HTML]{EFEFEF} 
    
    Ensemble Acc               &68.0   &52.9        &35.1       &56.3       &  &67.4        &49.5        &23.7       &48.0       \\ \hline 
    \rowcolor[HTML]{C1FFC1} 
    Ensemble ($\sigma$)                  &76.6   &62.9        &51.8       &60.2       &  &75.8        &61.5        &37.5   &53.1  \\ \hline
    \hline
                                & \multicolumn{9}{c}{SADE \cite{zhang2022SADE}}                                             \\ \cline{2-10} 
                            & \multicolumn{4}{c}{ImageNet-LT} &  & \multicolumn{4}{c}{CIFAR100-LT} \\ \cline{2-5} \cline{7-10} 
    \multirow{-3}{*}{Model} & Many   & Med.   & Few   & All   &  & Many   & Med.   & Few   & All   \\ \hline 
    
    E1 Acc                     &68.8   &43.7        &17.2       &49.8       &  &67.6        &36.3        & 6.8      &38.4       \\
    E2 Acc                     &65.5    &50.5        &33.3       &53.9       &  &61.2        &44.7        &23.5       &44.2       \\
    E3 Acc                     &43.4   &48.6        &53.9       & 47.3      &  &14.0        & 27.6       & 41.2      &25.8       \\
    \rowcolor[HTML]{EFEFEF} 
    Ensemble Acc               &67.0   &56.7        &42.6       &58.8       &  &61.6        &50.5        &33.9       &49.4       \\ \hline 
    \rowcolor[HTML]{C1FFC1} 
    Ensemble ($\sigma$)                &78.3   &62.4        &49.3       &61.4       &  &75.5        &57.1        &46.5       &59.8       \\ \hline
    \hline
                                    & \multicolumn{9}{c}{MDCS (ours)}                                             \\ \cline{2-10} 
                            & \multicolumn{4}{c}{ImageNet-LT} &  & \multicolumn{4}{c}{CIFAR100-LT} \\ \cline{2-5} \cline{7-10} 
    \multirow{-3}{*}{Model} & Many   & Med.   & Few   & All   &  & Many   & Med.   & Few   & All   \\ \hline 
    
    E1 Acc                     &   71.9 &  40.8      &  12.1     &  48.9      &  & 75.2       & 37.3       &4.1       &40.6       \\
    E2 Acc                     &    68.2    &    54.1    &  36.8     &  57.1    &  & 66.4      & 51.7       & 31.4      &  50.8     \\
    E3 Acc                     &    51.8    &    56.5    &  58.8     &  55.7     &  & 23.9       & 37.8       & 48.2      & 36.0      \\
    \rowcolor[HTML]{EFEFEF} 
    Ensemble Acc               &  72.6      &   58.1     &   44.3    &61.8        & &72.4        & 57.8       & 35.0      &  56.1     \\ \hline
    \rowcolor[HTML]{C1FFC1} 
    Ensemble ($\sigma$)               &81.2   &64.6        &53.4       &65.3       &  &81.8        &63.2        &55.2   &66.9 \\
    
    \bottomrule
   
    \end{tabular}
    
    }
    }
     \caption{Recognition accuracy (\%) and diversity factor (\%) of each expert and ensemble model in Many-shot, Medium-shot, and Few-shot categories. The experiment is conducted on CIFAR100-LT with IF = 100. The results show our method outperforms SOTA in terms of both expert diversity and a single expert's accuracy, which are critical to the performance of the ensemble model.}
    \label{lambda_compare}
   %\vspace{-10px}
\end{table}

\textbf{Comparison with SOTA methods.} The recognition accuracy and diversity factor results are shown in Table \ref{lambda_compare}, where we compare our result with the prior art, such as RIDE and SADE. The RIDE \cite{wang2020longRIDE} aims to improve diversity through KL-divergence between experts. However, simply maximizing the KL divergence between experts cannot lead to good diversity and accuracy. SADE is limited to only generating different inversely long-tailed data distributions by adjusting the hyper-parameter in the inverse softmax loss \cite{zhang2022SADE} and the accuracy and diversity of Many- or Medium-shot is severely inhibited for the "tail category expert," E3. The results show a significant advantage of our method in terms of both diversity and accuracy. The E1 trained with our DL shows significant improvement in diversity and accuracy in Many-shot categories and similar results in medium-shot categories and Few-shot categories. The E2 and E3 in MDCS also show great strengths in all three shots, which demonstrates  the effectiveness of our DL.

\textbf{The effect of $\lambda$ for diversity.}
Table \ref{diversity2} shows different $\lambda$ used in diversity loss to increase the model diversity. With $\lambda$ all set to 0, the experts focus on head or tail categories, which gives the model poor diversity. With $\lambda$ set to \{1, 1, 1\}, experts focus on average different categories, the model could get better diversity. With $\lambda$ set to \{0, 1, 2\} and \{-0.5, 1, 2.5\}, experts focus on different categories, the model gets best diversity.

\begin{table}[!htb]
	\centering
	\begin{tabular}{c|cccc}
    \toprule
    $\lambda$ &  Many  & Med. & Few  & All   \\
    \midrule
     \{0, 0, 0\}   &85.3  &59.2 &22.4  &55.4\\
     \{1, 1, 1\}   &80.6  &66.8 &47.4  &64.7\\
     \{2, 2, 2\}   &54.2  &58.4 &62.3  &58.2\\
     \{0, 0, 1\}   &84.4  &63.9 &37.6  &61.8\\
     \{1, 2, 2\}   &70.5  &63.3 & 60.8  &64.9\\
     \{0, 1, 2\}   &80.7  &64.4 & 51.2  &65.4\\
     \{-0.5, 1,2. 5\}   &\textbf{81.8}  &\textbf{63.2} &\textbf{55.2}  &\textbf{66.9}\\
    \bottomrule
\end{tabular}
    \caption{The effect $\lambda$ for the three-expert model on the CIFAR100-LT (IF = 100). Different combinations of $\lambda$ affect the diversity of the model.}
\label{diversity2}
 
\end{table}

\subsection{Lower Model Variance}

Model variance is the degree of variation in the predictions produced by the same model on different training datasets. With high model variance, the model may perform very differently on different training data, which may indicate that the model overfitted the training data and thus performs poorly in generalizing to unseen data. For n random data sets $\mathbb{B}_{(1)}$,..., $\mathbb{B}_{(m)}$, the k-th models trained on $\mathbb{B}_{(k)}$ will predict $y_{(k)}$ for instance $x$. The mean predicted value of these models is $\overline{y}$, which is denote:
\begin{equation} 
\begin{split}
    \overline{y} = \frac{1}{m} \sum_{k=1}^{m} y_{(k)}, 
\end{split}
\end{equation}

and the model variance denotes:

\begin{equation} 
\begin{split}
    \textbf{Var}(x, f) = \frac{1}{m} \sum_{k=1}^{m} (y_{(k)} - \overline{y})^2.
\end{split}
\end{equation}

To establish a benchmark for model variance, we compare our approach against three baseline methods: cRT \cite{LTkang2019decoupling}, RIDE \cite{wang2020longRIDE} and RIDE with label smoothing (LS) \cite{LS1}. These metrics are evaluated using twenty independently trained models, trained on CIFAR100-LT with 300 samples for class 0 (IF = 100) \cite{wang2020longRIDE}. In Table \ref{var1}, compare with cRT, RIDE and RIDE with LS, our model has better accuracy performance as well as lower model variance. It also suggests that our approach has better generalization than using ensemble model and label smoothing regularization to reduce model variance. 

\begin{table}[!htb]
	\centering
	\begin{tabular}{c|cccc}
    \toprule
      Method &  cRT   & RIDE & RIDE + LS & MDCS (Ours)   \\
    \midrule
     
     Var    &0.50  &0.42 &0.41 &0.36 \\
     Acc    &36.4  &40.5 &41.3 &46.1 \\
    \bottomrule
\end{tabular}
    \caption{Comparison of mean accuracy and variance of baselines and our MDCS based on CIFAR100-LT. The experiment settings follow RIDE \cite{wang2020longRIDE}.}
\label{var1}
%\vspace{-10px}
\end{table}

We also conduct experiments to show the effect of our proposed method on the model variance. Table \ref{var2} shows that the model with our consistency self-distillation (CS) effect reduces the model variance of each expert for the Many-, Medium-, and Few-shot subsets.
\begin{table}[!htb]
	\centering
	\begin{tabular}{c|cccc}
    \toprule
    method &  Many-shot & Medium-shot & Few-shot & All   \\
    \midrule
     w/o CS  &0.28  &0.42 &0.49  &0.40\\
     w/  CS  &0.24  &0.38 &0.46  &0.36 \\
    \bottomrule
\end{tabular}
    \caption{The effect of CS on the model variance for the three-expert model on the CIFAR100-LT (IF = 100).}
\label{var2}
 
\end{table}

\section{Experiments}
In this section, we perform experiments on five widely used datasets in long-tailed recognition, including CIFAR100/10-LT \cite{krizhevsky2009learning}, ImageNet-LT \cite{liu2019large}, Places-LT \cite{liu2019large}, and iNaturalist 2018 \cite{DBLP:journals/corr/HornASSAPB17}. After that, we conduct ablation experiments on the CIFAR100-LT and ImageNet-LT datasets to gain further insights.

\subsection{Dataset}

\textbf{CIFAR100/10-LT.}
CIFAR100/10-LT is the long-tailed version of CIFAR100/10 \cite{krizhevsky2009learning}. CIFAR-100/10 contains 50,000 images for training and 10,000 images for the validation of size 32 × 32 with 100/10 classes. Following \cite{wang2020longRIDE, zhang2021test}, we use the same long-tailed version for a fair comparison. The imbalanced factor (IF) $\beta$ is defined by $\beta$ = $N_{max}/N_{min}$, and this reflects the degree of imbalance in the data. The imbalance factors used in the experiment are set to 100 and 50.

\textbf{ImageNet-LT and Places-LT.}
ImageNet-LT and Places-LT are the long-tailed versions of the dataset ImageNet-2012 \cite{imagenet} and  the large-scale scene classification dataset Places \cite{zhou2017places} proposed by Liu \cite{liu2019large}. We follow their work by conducting the same dataset by sampling subsets following the Pareto distribution with the power value $\gamma = 6$. Overall, ImageNet-LT has 115.8K images from one thousand categories with an imbalanced factor $\beta = 1280/5.$ Places-LT contains 184.5K images from 365 categories with imbalanced factor $\beta=4980/5$.

\textbf{iNaturalist 2018.}
iNaturalist \cite{DBLP:journals/corr/HornASSAPB17} is a large-scale real-world dataset for long-tailed recognition, which suffers from extremely imbalanced distribution. It contains 437.5K training images and 24.4K validation images from 8142 categories. In addition, the fine-grained problem makes it more challenging \cite{8752297}. Moreover,  we follow the works \cite{li2022nested, zhang2022SADE} to divide classes into Many-shot(with more than 100 images), Medium-shot (with 20 - 100 images), and Few-shot (with less than 20 images) parts and report the results on each part.

\subsection{Implementation Details}
\textbf{Ensemble method.} The final ensemble is average across the experts.

\textbf{Architecture and settings.} We use the same setup for all the baselines and our method. Specifically, following previous work \cite{wang2020longRIDE,li2022nested,zhang2021test}, we employ ResNet-32 for CIFAR100/10-LT, ResNeXt-50/ResNet-50 for ImageNet-LT, ResNet-152 for Places-LT and ResNet-50 for iNaturalist 2018 as backbones, respectively. Moreover, we adopt the cosine classifier for prediction on all datasets. If not specified, we use the SGD optimizer with a momentum of 0.9 and set the initial learning rate as 0.1 with linear decay. We set $\lambda$ = \{-0.5, 1, 2.5\} and $\alpha=0.6$ in our method for all benchmarks. The results of our comparison method are taken from their original paper and our results are averaged over three experiments. More implementation details about epochs we have marked in the comparison table and the hyper-parameter statistics are reported in Appendix. 

\textbf{Augmentation.}
Our purposed CS utilizes weakly-augmented view and strongly-augmented view to conduct self-distillation. On the CIFAR10/100-LT dataset, the weak augmentation includes crop, horizontal flip, and rotation. The strong augmentation uses CIFAR10Policy besides the basic augmentation. For ImageNet-LT, Places-LT, and iNaturalist, we use cropping, horizontal flipping, rotation, and ColorJitter as weak augmentation. For a fair comparison, we utilize RandAug  \cite{cubuk2020randaugment} as strong augmentation for ImageNet-LT and iNaturalist 2018. We add RandomGrayscale and Gaussian Blur to the basic data augmentation composing strong augmentation for Places-LT following previous work \cite{li2022nested}. %The RandomGrayscale function randomly converts images to grayscale with a probability of 0.2, and the Gaussian Blur function chooses a number randomly from 0 to 2 as the radius of the filter kernel.

\subsection{Comparisons with SOTA on Benchmarks}
% We compare our purposed method MDCS with prior art including SADE\cite{zhang2022SADE}, BCL\cite{zhu2022balanced}, NCL\cite{li2022nested} and so on.

\begin{table}[!htb]
	\centering
	\begin{tabular}{c|p{1.0cm}p{1cm}|p{1cm}p{1cm}}
    \toprule
      Method & \multicolumn{2}{c}{CIFAR100-LT}  & \multicolumn{2}{c}{CIFAR10-LT} \\
      \midrule
      Imbalance Factor & 100   &  50  & 100 & 50\\
    \midrule
     200 epochs &~ & ~ & ~ & ~\\
     CB Focal loss \cite{reweightcao2019learning}    &39.6 &45.1 & 74.5& 79.2 \\
     LDAM+DRW \cite{reweightcao2019learning}       &42.0 &46.6 & 77.1 & 81.0  \\
     BBN\cite{LTzhou2020bbn}              &42.5 &47.0 & 79.8 & 81.1  \\
     LFME\cite{xiang2020learning}              &42.3 &- & - & -  \\
     CAM\cite{zhang2021bag}               &47.8 &51.7 & 80.0 & 83.6  \\
     Logit Adj.\cite{reweightmenon2020long}        &43.9 &-      & 77.7 & -  \\
     Hybrid-SC\cite{wang2021contrastive}         &46.7 &51.8 & 81.4 & 85.3 \\
     RIDE\cite{wang2020longRIDE}                    &49.1 &-      & -      & -  \\
     ResLT\cite{cui2022reslt}             &48.2 &52.7 & 82.4 & 85.3  \\
     SADE\cite{zhang2021test}              &49.8 &53.9 & - & -  \\
     BCL\dag\cite{zhu2022balanced}               &51.9 &56.5 & 84.3 & 87.2 \\
    \rowcolor[HTML]{FFFFCC} 
     MDCS\dag(Ours)        &\textbf{53.2} &\textbf{57.2} & \textbf{85.8} & \textbf{89.4}  \\
 \midrule
     400 epochs &~ & ~ & ~ & ~\\
     ACE\cite{cai2021ace} &49.6 &51.9  & 81.4 & 84.9   \\
     BSCE\dag\cite{ren2020balanced} &50.6 &55.0  & 84.0 & 85.8   \\
     PaCo\dag\cite{cui2021parametric} &52.0 &56.0  & - & -  \\
     SADE\dag\cite{zhang2021test} &52.2 &57.3  & - & -  \\
     NCL\dag\cite{li2022nested} &54.2 &58.2  & 85.5 & 87.3  \\
    \rowcolor[HTML]{FFFFCC} 
     MDCS\dag(Ours) &\textbf{56.1} &\textbf{60.1}  & \textbf{87.2} & \textbf{88.3}  \\

    \bottomrule
\end{tabular}
    \caption{Comparisons on CIFAR100-LT and CIFAR10-LT datasets with the IF of 100 and 50. \dag denotes models trained with RandAugment\cite{cubuk2020randaugment}.}
\label{CIFAR}
%\vspace{-10px}
\end{table}

\textbf{Long-Tailed CIFAR-100 and CIFAR-10.} The comparison results between MDCS and other methods on long-tailed CIFAR datasets are shown in Table \ref{CIFAR}. We conduct experiments on CIFAR100-LT and CIFAR10-LT with imbalance factors of 100 and 50. Additionally, for fairness, we compare results for 200 epochs and 400 epochs respectively. Our MDCS significantly outperforms the previous method on all groups, including 56.1\% on the CIFAR100-LT dataset with an IF of 100 when trained for 400 epochs. 

\textbf{ImageNet-LT, Places-LT, and iNaturalist 2018.}
Table \ref{tabImage}, \ref{tabPlaces}, and \ref{tabNat} list the Top-1 accuracy of SOTA methods utilizing different backbones on ImageNet-LT, Places-LT, and iNaturalist 2018, respectively. We report the overall Top-1 accuracy as well as the Top-1 accuracy on Many-shot, Medium-shot, and Few-shot groups for Place-LT, and iNaturalist 2018. For fair comparisons, we report the accuracy results at different epochs and these results are from their origin papers. Compared with prior arts, such as PaCo, BCL, NCL, and SADE, our proposed MDCS achieves SOTA performance in the same setting. %a better overall accuracy of 60.7\% with ResNet-50 and 61.8\% with ResNext-50, respectively.
 
\begin{table}[!htb]

	\centering
	\begin{tabular}{c|p{1cm}|cc}
    \toprule
      Method & multi-experts & ResNet-50  & ResNeXt-50 \\
 
    \midrule
     180 epochs & ~ & ~ & ~\\
     LADE\cite{hong2021disentangling} &~  &- & 53.0  \\
     BBN\cite{LTzhou2020bbn} &\Checkmark  &48.3 & 49.3  \\
     PaCo\dag\cite{cui2021parametric} &~  &- & 56.0  \\
     BCL\dag\cite{zhu2022balanced}  &~  &- & 57.1    \\
     SADE\cite{zhang2021test}  &\Checkmark  &- & 58.8  \\
 \midrule
 \midrule
     \rowcolor[HTML]{FFFFCC} 
     MDCS\dag(Ours) &\Checkmark &\textbf{59.3} & \textbf{60.2}    \\
     \midrule
     400 epochs &~ & ~ & ~  \\
     ACE\cite{cai2021ace}  &\Checkmark  &54.7 & 56.6    \\
     $\tau$-norm\cite{LTkang2019decoupling} &~  &54.5 & 56.0  \\
     BSCE\dag\cite{ren2020balanced} &~  &55.0 & 56.2 \\
     PaCo\dag\cite{cui2021parametric} &~  &57.0 & 58.2 \\
     NCL\dag\cite{li2022nested} &\Checkmark  &59.5 & 60.5    \\
 \midrule
 \midrule
     \rowcolor[HTML]{FFFFCC} 
     MDCS\dag(Ours) &\Checkmark &\textbf{60.7} & \textbf{61.8}    \\
 
    \bottomrule
\end{tabular}
\caption{Comparisons on ImageNet-LT. \dag denotes models trained with RandAug \cite{cubuk2020randaugment}.}
\label{tabImage}
%\vspace{-10px}
\end{table}

\begin{table}[!htb]
	\centering
	\begin{tabular}{c||ccc|c}
    \toprule
      Method  & Many & Medium & Few  & All \\

     OLTR\cite{liu2019large}   &44.7  & 37.0   & 25.3    &35.9\\
     $\tau$-norm\cite{LTkang2019decoupling}   &37.8  &40.7   &31.8    &37.9\\
     ResLT\cite{cui2022reslt}  &39.8  &43.6   &31.4    &39.8\\
     MiSLAS\cite{Zhong_2021_CVPR} &39.6  &43.3   &36.1    &40.4\\
     BSCE\dag\cite{ren2020balanced}   &- & -   & -    &40.2\\
     PaCo\dag\cite{cui2021parametric}   &36.1   & 47.9    &  35.3    & 41.2  \\
     NCL\dag\cite{li2022nested}    &-   & -    & -    & 41.8  \\
 \midrule
  \midrule
  \rowcolor[HTML]{FFFFCC} 
     MDCS\dag(Ours)  &43.1   & 42.9   & 36.3 & 42.4   \\
 
    \bottomrule
\end{tabular}
\caption{Comparisons on Places-LT, starting from an
ImageNet pre-trained ResNet-152 provided by Torchvision \cite{marcel2010torchvision}. \dag denotes models trained with RandAug \cite{cubuk2020randaugment}.}
\label{tabPlaces}
%\vspace{-10px}
\end{table}

\begin{table}[!htb]
	\centering
	\begin{tabular}{c||ccc|c}
    \toprule
      Method  & Many & Medium & Few  & All \\
      
    \midrule
     100 epochs &~ & ~ & ~ & ~\\
     BBN\cite{LTzhou2020bbn}    &49.4  & 70.8   & 65.3 & 66.3  \\
     $\tau$-norm\cite{LTkang2019decoupling} &65.6  & 65.3   & 65.9 & 65.2  \\
     BCL\dag\cite{zhu2022balanced}       & - & -  &- &71.8  \\
 \midrule
  \midrule
  \rowcolor[HTML]{FFFFCC} 
     MDCS\dag(Ours)   &71.8    & 73.1    & 72.4    & 72.5  \\ 
 
    \midrule
     200 epochs &~ & ~ & ~ & ~\\
     CE          &68.1 & 41.5   & 14.0& 48.2 \\
     RIDE\cite{wang2020longRIDE}            &70.5  &  73.7& 73.3 &73.2 \\
     SADE\cite{zhang2021test}             &74.5 & 72.5 & 73.0& 72.9 \\
   \midrule
    400 epochs &~ & ~ & ~ & ~\\
     ACE\cite{cai2021ace}  &- & -& - & 72.9   \\
     BSCE\dag\cite{ren2020balanced}   &72.3.  & 72.6.   & 71.7    & 71.8\\
     PaCo\dag\cite{cui2021parametric}   &70.3   & 73.2    & 73.6    & 73.2  \\
     SADE\dag\cite{zhang2021test}  &75.5    & 73.7    & 75.1    & 74.5  \\
     NCL\dag\cite{li2022nested}    &72.7   & 75.6    & 74.5    & 74.9  \\

 \midrule
  \midrule
  \rowcolor[HTML]{FFFFCC} 
     MDCS\dag(Ours)   &76.5    & 75.5    & 75.2    & 75.6  \\ 
 
    \bottomrule
\end{tabular}
\caption{Comparisons on iNaturalist 2018.  \dag denotes models trained with RandAugment\cite{cubuk2020randaugment}.}
\label{tabNat}
%vspace{-10px}
\end{table}

\section{Ablation Study and Further Analysis}
\label{Ablation_Study}

\begin{figure}[!htb]
\centering
\includegraphics[width=1\linewidth]{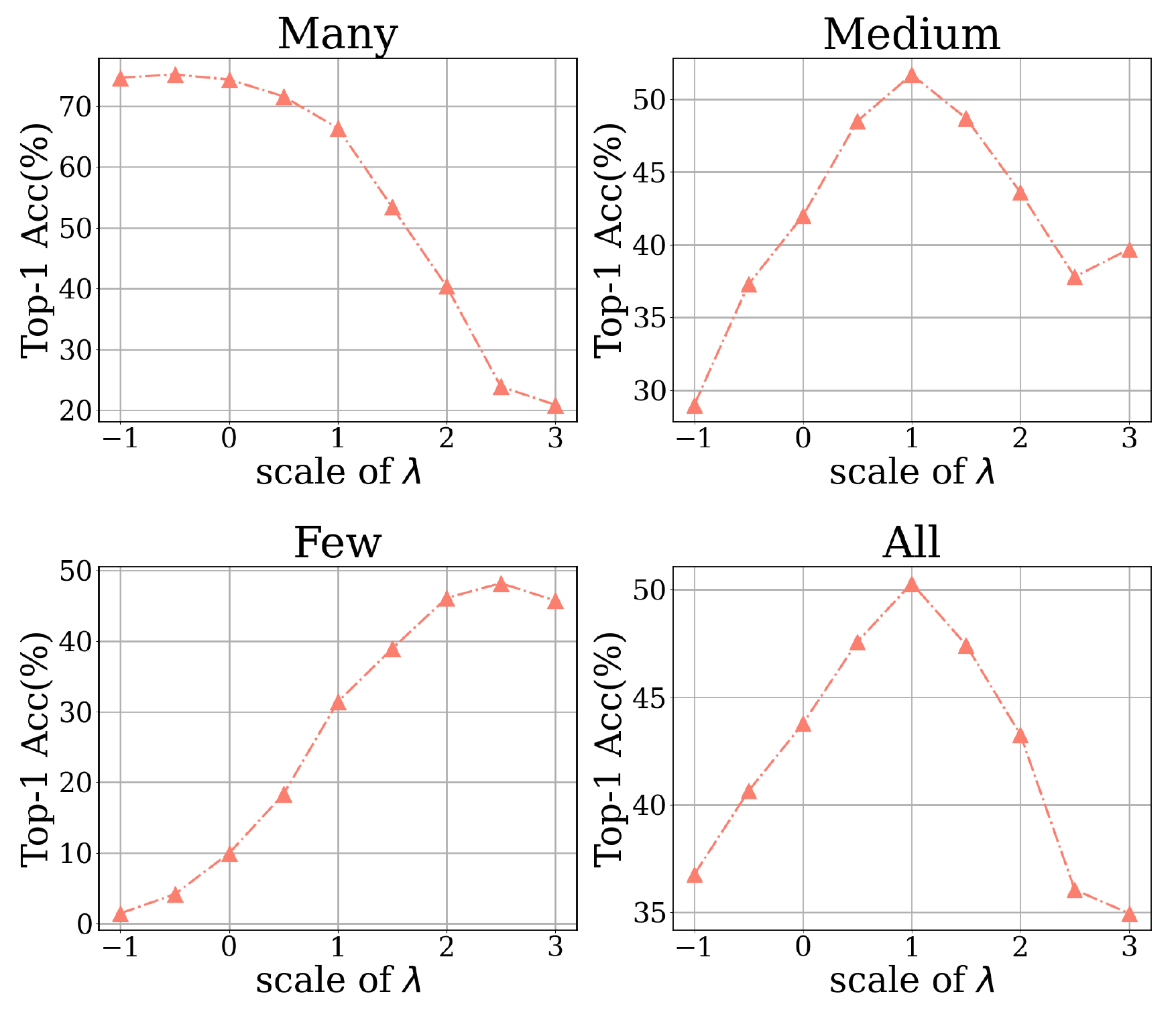}
\caption{Experiments on the effect of $\lambda$ to simulate weight distributions. The results demonstrate $\lambda$ has the ability to simulate different weight distributions focusing on Many-shot, Medium-shot, and Few-shot.}

\label{lambda}
%\vspace{-10px}
\end{figure}

\textbf{Simulating weight distributions by $\lambda$.}
To enrich the diversity of each expert, the Diversity Loss employs the hyper-parameter $\lambda$ to simulate different weight distributions for each expert's training. Fig. \ref{lambda} indicates how different $\lambda$ can affect the accuracy of Many-shot, medium-shot, and few-shot categories and overall accuracy. When $\lambda$ increases, the accuracy of Many-shot categories decreases while the accuracy of few-shot categories increases, which demonstrates the ability to simulate different weight distributions. Besides, when $\lambda$ gets high enough, the accuracy of few-shot classes will decrease. This is due to the few-shot group having 30 categories on CIFAR100-LT, this categorization is not fine-grained enough to cater to the effect of the extremely inversely long-tailed distribution generated.

\textbf{Influence of data augmentations.}
The RandAug  \cite{cubuk2020randaugment} is widely employed as its strong generalization for long-tailed recognition \cite{li2022nested, zhu2022balanced, cui2021parametric}. In this subsection, we conduct different augmentations on training samples to evaluate the effectiveness of weak-strong consistency self-distillation. The results are shown in Table \ref{tab:self-distillation}, where we compare weak-strong distillation with weak-weak distillation and strong-strong distillation. The results of weak-strong consistency self-distillation are better than strong-strong distillation, which demonstrates that the excellence of our structure does not depend entirely on RandAugment.

\begin{figure}[!htb]
\centering

\includegraphics[width=1\linewidth]{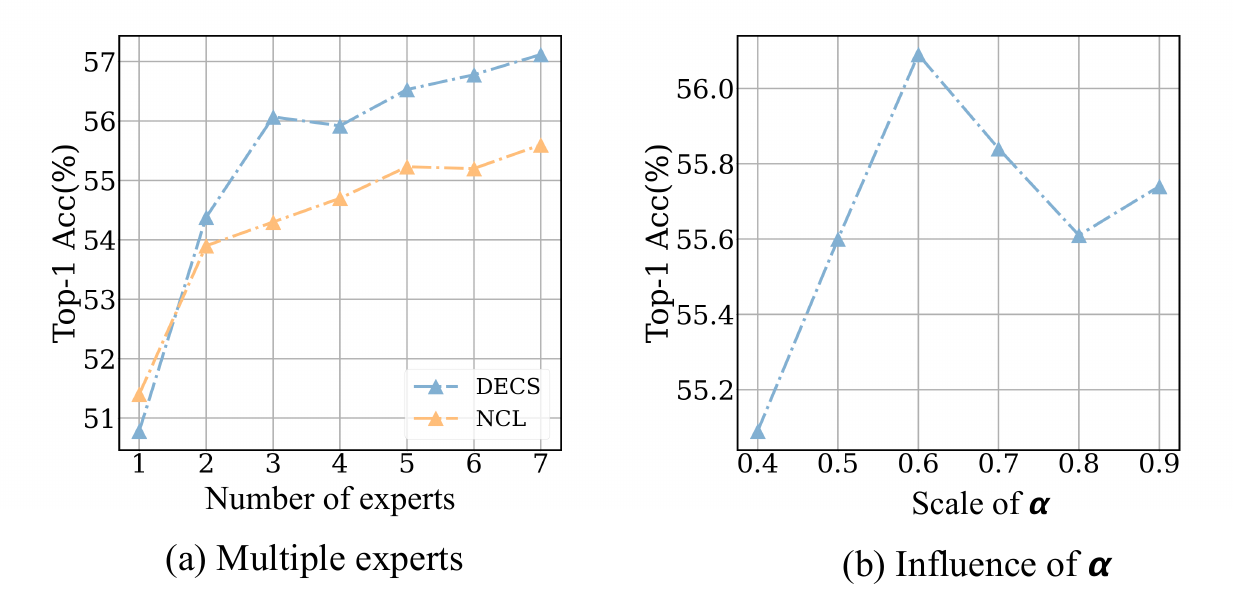}
\caption{(a) Comparison of using different number of experts with NCL \cite{li2022nested}. We report the performance of all categories. When the number of experts increases, the model's performance also tends to improve. (b) the loss of weight of Consistency Self-distillation. The best result is achieved when $\alpha = 0.6$.  }

\label{multi}
%\vspace{-10px}
\end{figure}

\begin{table}[!htb]
  \centering
  \begin{tabular}{cc|c}
    \toprule
    View1 & View2 & Top1-Acc\\
    \midrule
    Weak augmentation& Weak augmentation&  50.7\\
    Strong augmentation& Strong  augmentation& 54.6\\
    Weak augmentation& Strong augmentation & 56.1\\
    \bottomrule
  \end{tabular}
  \caption{Comparisons of training the model with weak-weak augmentation, strong-strong augmentation, and weak-strong augmentation. Experiments are conducted on CIFAR100-LT with IF = 100. }
  \label{tab:self-distillation}
  %\vspace{-10px}
\end{table}
\textbf{Impact of a different number of experts.}
Our proposed MDCS is a multi-expert framework. The number of experts can be easily extended by adjusting $\lambda$ for different experts. We conduct experiments to demonstrate the power of multiple experts. As shown in Fig. \ref{multi} (a), the performance of MDCS tends to get better when the number of experts increases. The $\lambda$ for the one-expert model is $\{1\}$, for the two-expert model is $\{-0.5, 2.5\}$, for the three-expert model is $\{-0.5, 1, 2.5\}$, for the four-expert model is $\{-0.5, 0, 1, 2.5\}$, for the five-expert model is $\{-0.5, 0, 1, 2, 2.5\}$, for the six-expert model is $\{-1, -0.5, 0, 2, 2.5, 3\}$, and for the seven-expert model is $\{-1, -0.5, 0, 1, 2, 2.5, 3\}$.

\textbf{The rule for setting $\lambda$.}
The ensemble models are not sensitive to the  hyper-parameter $\lambda$ within a reasonable range, so we can easily choose $\lambda$ just to spread across this range. When the number of branches of experts increases, we first average divide experts into the three groups to set $\lambda$. For the head group, the $\lambda \in$ [-1, 0.5], for the balanced group, the $\lambda \in$ (0.5, 1.5), for the tail group, the $\lambda \in$ [1.5, 3]. when the values of $\lambda$ for different experts fell within the above three ranges, the multi-expert model exhibits effective performance improvements. For example, we set $\{-0.5, 1, 2.5\}$ for three experts and $\{-1, -0.5, 0, 1, 2, 2.5, 3\}$ for seven experts.

\textbf{Influence of loss weight $\alpha$.}
The $\alpha$ is an adjusted loss weight of Consistency Self-distillation to control the contribution of the CS part in total loss. To find an appropriate for $\alpha$, A series of values experimented on the CIFAR100-LT dataset. As shown in Fig. \ref{multi} (b), the best performance is achieved when $\alpha=0.6$. The best result means a balance of supervised learning and self-distillation. 
% $\alpha=0.4$ 55.02,
% $\alpha=0.5$ 55.64,
% $\alpha=0.6$ 55.87,
% $\alpha=0.7$ 55.09.

\textbf{Ablation studies on all components.}
\begin{table}[!htb]
  \centering
  \begin{tabular}{ccc|c}
    \toprule
     DL &  w/RandAug & CS  & Accuracy(\%) \\
    \midrule
    \XSolidBrush &\XSolidBrush & \XSolidBrush   & 47.8\\
    \XSolidBrush &\Checkmark & \Checkmark     & 48.5\\
    \Checkmark &\XSolidBrush & \XSolidBrush    & 50.7\\
    \Checkmark   &\Checkmark & \XSolidBrush     & 54.1\\
    %\Checkmark   &\Checkmark & Normal       & 54.6\\
    \Checkmark   &\Checkmark & \Checkmark      & 56.1\\
    \bottomrule
  \end{tabular}
  \caption{Ablation study on the CIFAR100-LT dataset with an IF of 100. The DL indicates Diverse Loss. w/RandAug means weak-strong augmentation. The \XSolidBrush in w/RandAug means conducting weak-weak augmentation. }
  \label{all}
  %\vspace{-15px}
\end{table}
%\vspace{-10px}
In this subsection, we conduct detailed ablation studies on the CIFAR100-LT dataset to analyze every component of our MDCS. As shown in Table \ref{all}, we evaluate the proposed components including Diverse Loss (DL), weak-strong augmentation(w/RandAug), Consistency Self-distillation (CS), respectively. The \XSolidBrush  in DL means we use normal Softmax to conduct experiments and the \XSolidBrush in w/RandAug means we employ weak-weak augmentation. As shown in Table, our proposed Diversity Loss improves the performance by 2.9\%. It is a core component because, without our DL, the other components are less effective at improving performance. MDCS Employing weak-strong augmentation can improve performance from 50.7\% to 54.1\%, which proves the strength of RandAug \cite{cubuk2020randaugment, li2022nested}. Eventually, when conducting our proposed CS, the performance is significantly further improved, from 54.6\% to 56.1\%.
  %\vspace{-10px}

\section{Conclusion}

In this paper, we propose a novel method, MDCS, to cater to the diversity and variance of multi-expert, leading to improved long-tailed recognition accuracy. Our MDCS contains two core components: (1) diversity loss (DL), which can effectively enhance the diversity of experts. (2) consistency self-distillation (CS), which is a novel self-distillation method for reducing the model variance. Furthermore, we propose confident instance sampling in CS to ensure unbiased knowledge. In analyses and ablation studies, we analyze the effectiveness of our proposed core components through experimental results. Moreover, the roles of our DL and CS are mutually reinforcing and coupled. Experimental evidence shows that our MDCS achieves significant performance over the SOTA methods on five popular benchmarks, including 56.1\% (+1.9\%) accuracy on CIFAR100-LT with an IF 100, 61.8\% (+1.3\%) accuracy on ImageNet-LT with ResNeXt-50, and 75.6\% (+0.7\%) accuracy on iNaturalist 2018 with ResNet-50.

\section*{Acknowledgement}
This work was supported in part by the National Natural Science Foundation of China under Grant No.62271034, and in part by the Fundamental Research Funds for the Central Universities under Grant XK2020-03.

\section*{Appendix}
\textbf{1. The Efficiency of Consistency Self-distillation}

\begin{figure}[!ht]
\centering
\includegraphics[width=1\columnwidth]{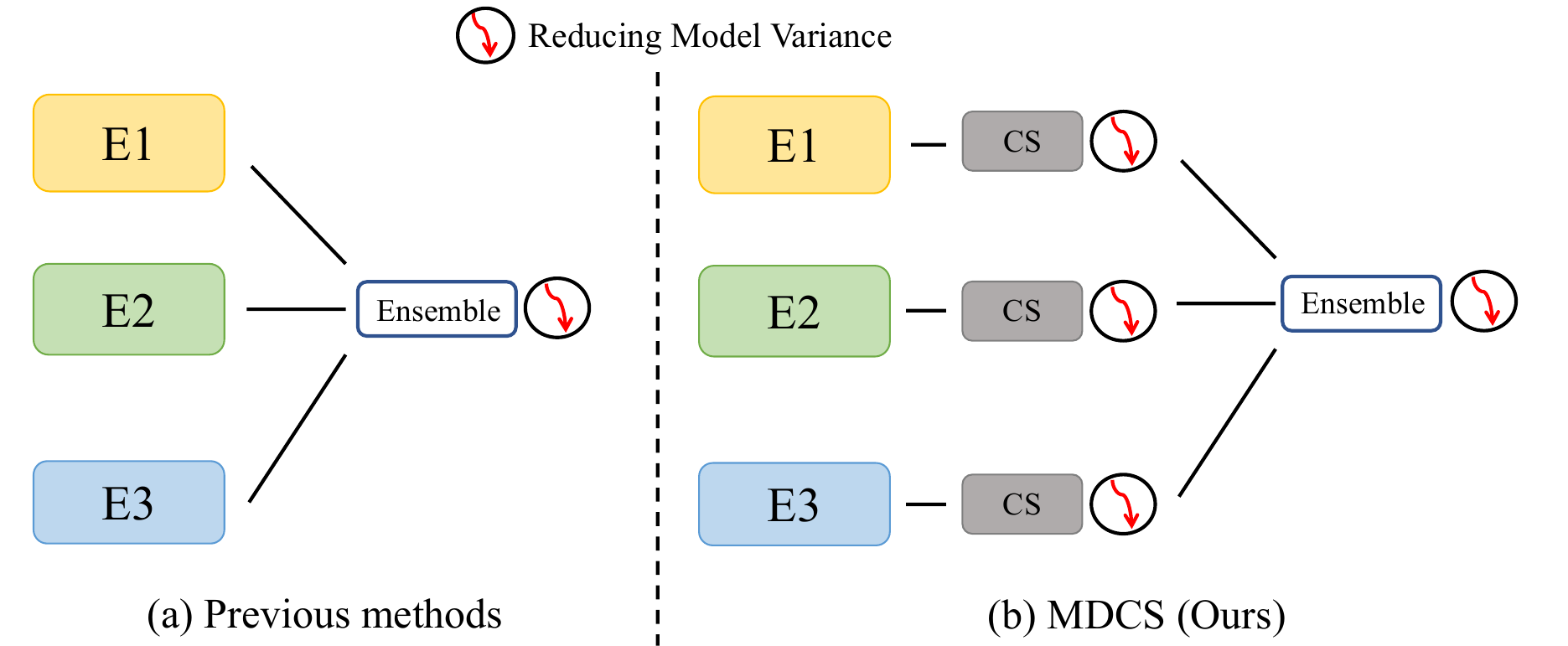}
\caption{}

\label{sup1}
%\vspace{-10px}
\end{figure}

As illustrated in Fig. \ref{sup1}, previous methods \cite{wang2020longRIDE, cai2021ace, zhang2022SADE} reduced the model variance only by using an ensemble of multiple experts. In contrast, our approach not only reduces the variance by ensemble but also reduces the model variance by CS for each expert. The effect of CS is not only to reduce the model variance. Each expert gets richer constraint information through weakly augmented images, which enhances the expert's own recognition ability. As shown in Table \ref{cs}, experts with stronger recognition abilities also produce more diverse ensemble models. 

\begin{table*}[!htb]%\small
	\centering
	\begin{tabular}{c|ccccc}
    \toprule
      Method &  E1 Acc & E2 Acc & E3 Acc & All Acc & $\sigma$   \\
    \midrule
     w/o CS  &38.8  &45.2 &31.4  &50.7 &53.4\\
     w/ CS    &40.6\textcolor{blue}{(+1.8)}&50.8\textcolor{blue}{(+5.6)}  &36.0\textcolor{blue}{(+4.6)} &56.1\textcolor{blue}{(+5.4)} &60.4\textcolor{blue}{(+7.0)} \\
     
    \bottomrule
\end{tabular}
    \caption{The efficiency of Consistency Self-distillation. With CS, not only is the model variance reduced, but also the expert recognition ability and the final model diversity are improved.}
\label{cs}
\vspace{-10px}
\end{table*}

\begin{table*}[!t]
\centering
\begin{tabular}{ccccc}
\hline
\multicolumn{1}{c|}{Items}               & \multicolumn{1}{c|}{CIFAR100/10-LT}   & \multicolumn{1}{c|}{ImageNet-LT} & \multicolumn{1}{c|}{Places-LT}    & iNaturalist 2018     \\ \hline
\multicolumn{5}{c}{Network Architectures}                                                                                                                                  \\ \hline
\multicolumn{1}{c|}{network backbone}    & \multicolumn{1}{c|}{ResNet-32}    & \multicolumn{1}{c|}{ResNeXt-50/ResNet-50}  & \multicolumn{1}{c|}{ResNet-152}   & ResNet-50            \\ \hline
\multicolumn{5}{c}{Training Phase}                                                                                                                                         \\ \hline
\multicolumn{1}{c|}{epochs}              & \multicolumn{1}{c|}{200/400}          & \multicolumn{1}{c|}{180/400}            & \multicolumn{1}{c|}{30}           &    \multicolumn{1}{c|}{100/400}                    \\ \hline
\multicolumn{1}{c|}{batch size}          & \multicolumn{1}{c|}{64}           & \multicolumn{1}{c|}{256}            & \multicolumn{1}{c|}{64}           &   \multicolumn{1}{c|}{512}                   \\ \hline
\multicolumn{1}{c|}{learning rate (lr)}  & \multicolumn{1}{c|}{0.1}          & \multicolumn{1}{c|}{0.1}            & \multicolumn{1}{c|}{0.01}         &  \multicolumn{1}{c|}{0.2}                    \\ \hline
\multicolumn{1}{c|}{lr schedule}         & \multicolumn{1}{c|}{linear decay}  & \multicolumn{1}{c|}{cosine decay}            & \multicolumn{1}{c|}{linear decay}  &    \multicolumn{1}{c|}{linear decay}                  \\ \hline
\multicolumn{1}{c|}{$\lambda$}                   & \multicolumn{1}{c|}{-0.5, 1, 2.5} & \multicolumn{1}{c|}{-0.5, 1, 2.5}            & \multicolumn{1}{c|}{-0.5, 1, 2.5} & \multicolumn{1}{c|}{-0.5, 1, 2.5}                       \\ \hline
\multicolumn{1}{c|}{weight decay factor} & \multicolumn{1}{c|}{$5 * 10^{-4}$}     & \multicolumn{1}{c|}{$5 * 10^{-4}$}            & \multicolumn{1}{c|}{$5 * 10^{-4}$}     &  \multicolumn{1}{c|}{$5 * 10^{-4}$}                     \\ \hline
\multicolumn{1}{c|}{momentum factor}     & \multicolumn{4}{c}{0.9}                                                                                                         \\ \hline
\multicolumn{1}{c|}{optimizer}           & \multicolumn{4}{c}{SGD optimizer with nesterov}                                                                                 \\ \hline
\end{tabular}
\caption{Statistics of the used network architectures and hyper-parameters in our experiments.}
\label{tabNat}
\end{table*}

\textbf{2. More Details Settings}

We implement our method with PyTorch. Following \cite{zhang2022SADE,li2022nested}, we use ResNeXt-50/ResNet-50 for ImageNet-LT, ResNet-32 for CIFAR100/10-LT, ResNet-152 for Places-LT and ResNet-50 for iNaturalist 2018 as backbones, respectively. Moreover, we adopt the cosine classifier for prediction on all datasets. The details settings for our method are shown in table \ref{tabNat}.

{\small
\bibliographystyle{ieee_fullname}
\bibliography{egbib}
}

\end{document}